\documentclass[utf8]{FrontiersinVancouver} 
\usepackage{url,hyperref,lineno,microtype,subcaption}
\usepackage[onehalfspacing]{setspace}

\usepackage{cite}
\usepackage{amsmath,amssymb,amsfonts}
\usepackage{algorithmic}
\usepackage{graphicx}
\usepackage{textcomp}
\usepackage{tabularx}
\usepackage{makecell}
\usepackage{multirow}
\usepackage{booktabs}

\def\keyFont{\fontsize{8}{11}\helveticabold }

\def\firstAuthorLast{Jason Jordan {et~al.}} 
\def\Authors{Jason Jordan\,$^{1}$, Mohammadreza Akbari Lor\,$^{1}$, Peter Koulen\,$^{2,*}$ Mei-Ling Shyu\,$^{1}$ and Shu-Ching Chen\,$^{3}$, }

\def\Address{$^{1}$School of Science and Engineering, University of Missouri-Kansas City, Kansas City, MO, USA \\
$^{2}$Vision Research Center, Department of Ophthalmology, School of Medicine, University of Missouri-Kansas City, Kansas City, MO, USA  \\
$^{3}$Data Science and Analytics Innovation Center, University of Missouri-Kansas City, Kansas City, MO, USA}

\def\corrAuthor{Peter Koulen}

\def\corrEmail{koulenp@umkc.edu}

\begin{document}

\onecolumn
\firstpage{1}

\title[MDF-MLLM: Deep Fusion Through Cross-Modal Feature Alignment]{MDF-MLLM: Deep Fusion Through Cross-Modal Feature Alignment for Contextually Aware Fundoscopic Image Classification}

\author[\firstAuthorLast ]{\Authors} 
\address{} 
\correspondance{} 

\extraAuth{}

\maketitle

\begin{abstract}

\textbf{Introduction:} This study aimed to enhance disease classification accuracy from retinal fundus images by integrating fine-grained image features and global textual context using a novel multimodal deep learning architecture. Existing multimodal large language models (MLLMs) often struggle to capture low-level spatial details critical for diagnosing retinal diseases such as glaucoma, diabetic retinopathy, and retinitis pigmentosa.

\textbf{Methods:} This model development and validation study was conducted on 1,305 fundus image–text pairs compiled from three public datasets (FIVES (n=800), HRF (n=45), and StoneRounds.org (n=460)), covering acquired (e.g., diabetic retinopathy, glaucoma), and inherited (e.g., retinitis pigmentosa, Stargardt) retinal diseases, and evaluated using classification accuracy and f1-score. The MDF-MLLM integrates skip features from four U-Net encoder layers (with ResNet50 backbone) into cross-attention blocks (layers 3, 18, 28, 38) within a LLaMA 3.2 11B MLLM. Vision features are patch-wise projected and fused using scaled cross-attention and FiLM-based U-Net modulation. Fusion and baseline models were trained on a distributed 12-GPU system. 

\textbf{Results:} Baseline MLLM achieved 60\% accuracy on the dual-type disease classification task. MDF-MLLM, with both U-Net and MLLM components fully fine-tuned during training, achieved a significantly higher accuracy of 94\%, representing a 56\% improvement. Recall and F1-scores improved by as much as 67\% and 35\% over baseline, respectively. Freezing the U-Net or MLLM during fusion demonstrated inferior performance (37\%–78\%). Ablation studies confirmed that the multi-depth fusion approach contributed to substantial gains in spatial reasoning and classification, particularly for inherited diseases with rich clinical text. Examples included correctly attributing XL Retinitis Pigmentosa based on fundus appearance and genetic metadata, whereas the baseline MLLM frequently hallucinated or misclassified similar cases. The model reached optimal U-Net performance after 18 epochs but required 55 epochs for full MLLM convergence, revealing a training-speed mismatch as a critical optimization constraint.

\textbf{Discussion:} MDF-MLLM presents a generalizable, interpretable, and modular framework for fundus image classification, outperforming traditional MLLM baselines through multi-scale feature fusion. The architecture holds promise for real-world deployment in clinical decision support systems. Future work will explore synchronized training techniques, a larger pool of diseases for more generalizability, and extending the model for segmentation tasks.

\tiny
 \keyFont{ \section{Keywords:} Cross-modal fusion,
Multimodal large language model,
U-Net,
Retinal fundus photography,
Vision transformer
}

\end{abstract}

\section{Introduction}
Ophthalmologists rely heavily on medical imaging to help discern key features within the human eye. Fundus photography is especially important, as it gives a detailed and colored view of the back of the eye, highlighting the structural components of the retina and its blood vessel pattern. The shape and structure of these components provide ophthalmologists and retina specialists with the necessary information for detecting abnormalities associated with eye diseases such as glaucoma \citep{LEE202215}, diabetes \citep{Sabanayagam2015}, age-related macular degeneration \citep{Deng2022-kl}, and hypertension \citep{Chew2012-sp}. The vascular structure of the retina is of particular importance, with retinal blood vessel shape, size, and damage all serving as important indicators of vision health, as well as overall cardiovascular health \citep{AboDan2025, Poplin2018-be}. Therefore, the development of diagnostic and data analysis tools that can provide a multi-depth understanding of local and global features to efficiently detect diagnostically relevant retinal structures would be very useful for ophthalmologists and retina specialists. With the rapid advancement of deep learning methods, large language models (LLMs), and other forms of artificial intelligence (AI), it appears necessary that for future biomedical research to provide broad clinical impact, it needs to generate systems that allow multiple AI model architectures to collaborate with each other, instead of in isolation, to advance their overall capabilities. Here, such an architecture is presented that enables the acquisition of multimodal (image and text) biomedical knowledge and provides helpful insights for medical practitioners within clinical settings.

Recent advances in large language models (LLMs), particularly those optimized for multimodal integration, open new engineering possibilities for designing systems that can combine visual and textual data in a unified framework. These multimodal LLMs (MLLMs) offer a pathway toward more context-aware diagnostic support tools by enabling information fusion across data types. However, challenges remain in adapting general-purpose models to specialized clinical domains. To some extent, MLLM architectures have adapted to the increasing diversity in information sources but have continued to struggle to grasp low-level visual perception. These limitations of perception result in poor recognition of smaller objects and nuances within images \citep{zhang2024exploring}. In contrast, other research has sought to improve this low-level understanding by incorporating MLLMs with segmentation models that harness the informational context in language outputs to effectively label (pixel-wise) features in the original image to create a user-enhanced visual understanding of the MLLM outputs \citep{lai2023lisa, yang2024empoweringsegmentationabilitymultimodal}. While these demonstrate respectable performance, they focus on the utilization of an existing (out-of-the-box) MLLM pre-trained for generalized knowledge and adapt the base outputs to expand the representations within the same general domain.

This study sought to expand on these concepts by introducing MDF-MLLM (Multi-Depth Fusion of MLLM), a cross-modal integration architecture that combines fine-grained image features with high-level contextual embeddings to improve the classification capability of an open-source MLLM (LLaMA 3.2 11B Vision-Instruct, \href{https://ai.meta.com/blog/llama-3-2-connect-2024-vision-edge-mobile-devices/}{ai.meta.com}) in medical imaging diagnostics. The MDF-MLLM architecture integrates the embeddings of the interior layers with the embeddings of the encoder from a classical U-Net segmentation model through customized fusion blocks. The proposed fusion methods optimize the MLLM model for image classification, which was demonstrated on open-source retinal fundus datasets (FIVES \citep{Jin2022-kq}, HRF \citep{budaiHRF2013}, and the Disease Atlas from StoneRounds.org \citep{stoneroundsWebsite}) categorized across distinct disease classes. Fundus photography is a cornerstone of ophthalmologic diagnostics and offers high-resolution, colorized visualization of the retinal surface and vasculature, enabling the identification of structural changes linked to various diseases. The main contributions of the research presented here can be summarized as follows:

1.	Demonstrated the capability of a multimodal LLM to integrate and optimize contextual visual information from individual U-Net model layers through enhanced fusion techniques that incorporate semantic and spatial alignment.

2.	Assessed the effectiveness of facilitating transfer learning on model performance through supplementary fine-tuning on open-source and customized retinal fundus image datasets.

3.	Validated the multimodal LLM's ability to harness the appropriate contextual information and effectively enhance predictive capacity for disease category classification from retinal fundus images and patient clinical text information while presenting the response predictions in a clinically relevant fashion for effective reasoning and interpretability.

The remainder of this paper provides an overview of related work within the fields of ophthalmology and AI in Section~\ref{sec:relwork}. Section~\ref{sec:methods} outlines the base model architecture and provides a theoretical framework for the proposed feature fusion methods that focuses on the enhancement of the multimodal LLM functionality. Section~\ref{sec:expresults} describes the dataset structure, provides a description of the experimental setup with evaluation metrics for overall performance assessment, and highlights the key findings of the study. The concluding remarks in Section~\ref{sec:conclusion} discuss the unique contributions of this research and offer additional insights into future research opportunities.

\section{Related Work}
\label{sec:relwork}
In medicine and, more specifically, ophthalmology, AI/ML has progressed along three largely independent tracks: (1) pixel-level segmentation, which extracts the anatomic priors clinicians need; (2) large language models (LLMs) that parse text notes or generate narrative reports; and (3) vision–language fusion frameworks that promise end-to-end clinical reasoning but are still in their infancy for retinal care. This section reviews each track, highlighting how recent advances motivate our cross-modal feature-learning approach.

\subsection{Deep Learning in Retinal Vessel Segmentation}
\label{subsec:deepl}
Early convolutional pipelines framed vessel extraction as an edge-finding task. DeepVessel, for example, learned hierarchical representations that delineated vessel boundaries and already exceeded handcrafted filters on DRIVE (Digital Retinal Images for Vessel Extraction)  \citep{staal2004ridge} and STARE (Structured Analysis of the Retina) \citep{hoover2000stare} benchmark datasets by a large margin, particularly on small diameter capillaries \citep{deepvessel2016}. U-Net, however, utilizes convolutional networks to extract low-level features from an image by utilizing a contracting encoder to capture context, followed by a symmetric expanding decoder path to enable precise localization, and introduced encoder–decoder skip connections that became the standard for ophthalmic segmentation \citep{10.1007/978-3-319-24574-4_28, 10643318}. Many variants of this system have been developed and stem from variations in the base convolutional network. One such architecture is Deformable-U-Net (DU-Net), which enriched U-Net with deformable convolutions to adaptively adjust for the vessels' shapes and sizes \citep{JIN2019149}. Dense Residual Vessel U-Net (DR-VNet) is another that utilizes residual dense net modules along with squeeze-and-excitation blocks for improving the task with small-diameter blood vessels \citep{10.1007/978-3-031-09037-0_17}.

Later, attention mechanisms were also utilized. For example, SA-U-Net (Spatial Attention U-Net) applied spatial-attention gates explicitly re-weighting the information in the final encoding layer before the decoding steps \citep{9413346}. MSMA-Net (Multi-Scale Multi-Attention Network) fuses multi-scale squeeze–excitation with attention to boost accuracy \citep{Kande2025-jx}, and Galdrán et al. showed that carefully tuned “minimal” U-Net architectures can rival heavier models on public benchmarks \citep{Galdran2022-wo}. Adversarial architectures have also been recently utilized. For example, SEGAN (Symmetric Equilibrium Generative Adversarial Network) casts segmentation as an adversarial game; its generator sharpened vessel edges while the discriminator enforced global coherence, increasing F1-scores on DRIVE, STARE and CHASEDB1 (Child Heart and Health Study \citep{6224174}) datasets \citep{ZHOU2021118}.

\subsection{Large Language Models in Ophthalmology}

As LLMs gain traction in clinical medicine, their application in ophthalmology is emerging as a key area of interest, prompting early evaluations of their utility and risk \citep{10.3389/fopht.2024.1387190, tabuse2025evaluatinglargelanguagemodels, Grzybowski2024-nn}. A recent review on the integration of LLMs in ophthalmology identified more than 100 relevant studies, highlighting promising applications in triage, clinical documentation, and patient education \citep{10.3389/fmed.2023.1291404}. However, the review also emphasized key limitations, including hallucinations, the spread of misinformation, concerns about medical liability, and threats to scientific integrity. Another study demonstrated that ChatGPT improved patient satisfaction by generating well-structured and coherent responses to various topics, including common retinal diseases \citep{potapenko2023artificial}. Despite these early successes, practical and ethical challenges must be addressed before LLMs can be reliably implemented in real-world clinical settings \citep{Betzler2023-bv}. Moreover, as vision capabilities in generative AI are still in the early stages of development, widespread adoption in ophthalmology remains a long-term prospect \citep{Sonmez1335, FENG2024100090}.

\subsection{Cross-Modal Fusion}
General-purpose vision-language pre-training offers mechanisms to close the previously mentioned gap surrounding the limitations of LLMs in highly specialized domains. BLIP-2 (Bootstrapping Language-Image Pre-training) bridges frozen image encoders and frozen LLMs with a lightweight Querying Transformer \citep{pmlr-v202-li23q}. The Flamingo architecture extends this idea to arbitrarily interleaved image–text sequences, enabling few-shot learning in captioning, visual question answering (VQA), and video tasks \citep{NEURIPS2022_960a172b}. Beyond single-image tasks, FFA-GPT (Fundus Fluorescein Angiography-Generative Pre-trained Transformer) \citep{Chen2024-uu} couples a BLIP \citep{pmlr-v162-li22n} vision encoder with a LLaMA \citep{DBLP:journals/corr/abs-2302-13971} decoder to auto-generate fluorescein-angiography reports and answer follow-up questions. DeepDR-LLM (Deep learning Diabetic Retinopathy) integrates a DeepDR-Transformer and a domain-fine-tuned LLaMA to output individualized management recommendations for patients and severity grading of diseases \citep{Li2024-gb}.

Medical-specific adaptations of MLLM architectures are also emerging. The Cross-Modal Augmented Transformer (CAT) injects attention blocks between retinal encoders and language decoders to generate more coherent diagnostic reports \citep{10857391}. Despite these advances, several critical limitations remain. Most frameworks (e.g., BLIP-2, Flamingo) rely heavily on pretrained, general-purpose encoders, which often fail to capture fine-grained spatial or structural features such as vessel morphology or micro-lesions that are crucial in ophthalmic diagnostics. Also, while models like FFA-GPT and DeepDR-LLM demonstrate task-specific utility, they are designed and fine-tuned around narrow clinical use cases and lack the flexibility to generalize across broader retinal disease categories. Finally, surveys highlight that current MLLMs often struggle to ground detailed visual attributes in the absence of explicit textual annotations, leading to poor sensitivity to subtle but clinically important findings \citep{MIHALACHE2025100154}. From a system's angle, the current literature on deep model fusion highlights practical challenges like parameter misalignment, computational cost, and catastrophic interference, when merging large vision and language backbones in a single deployable model \citep{li2023deepmodelfusionsurvey}.

Therefore, there is a clear gap in developing architectures that can simultaneously preserve low-level semantic precision from medical images and leverage the high-level contextual reasoning power of MLLMs in a unified model architecture. Our proposed MDF-MLLM addresses this by integrating U-Net–based multi-scale features with LLaMA’s vision–language embeddings through multi-depth cross-attention fusion, thereby enhancing both spatial grounding and clinical interpretability.

\section{Materials and Methods}
\label{sec:methods}
\begin{figure*}[h]
  \centering
  \includegraphics[width=\linewidth]{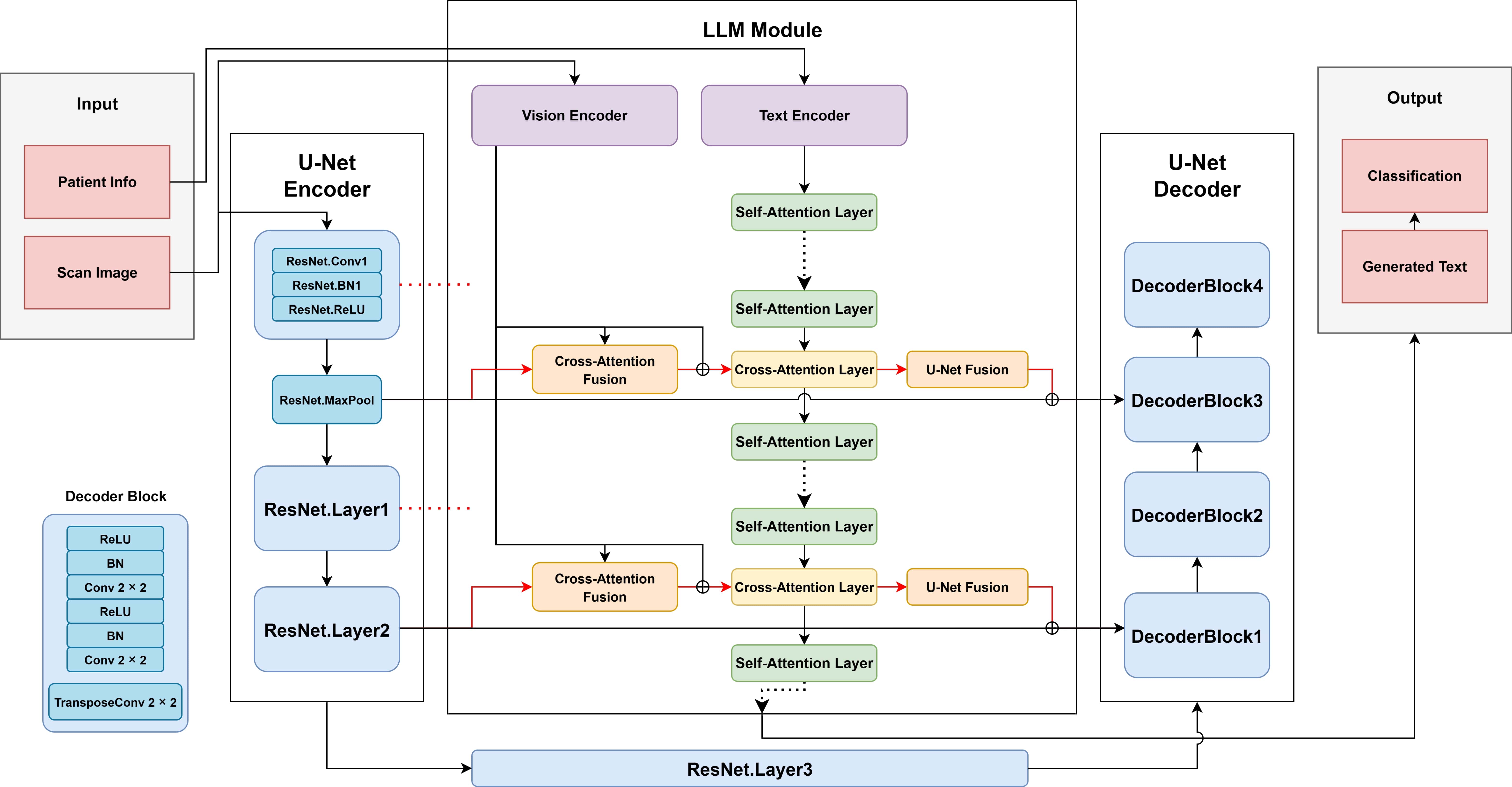}
  \caption{Overall architecture MDF-MLLM, showing U-Net model with ResNet50 encoder backbone integrated with LLaMA 3.2 11B Vision model cross-attention layers via customized fusion modules. The black arrows are the existing connections of the U-Net and LLM model, while the red arrows symbolize the connections made for the new fusion components. Dotted lines symbolize replicated fusion block pipelines processed through additional connection layers in the MLLM text encoder, and have been omitted for clarity. Detailed internal design of the cross-attention fusion blocks and U-Net decoder blocks is also presented above the overall architecture for better visual clarity.}
  \label{fig:overall}
\end{figure*}

The proposed MDF-MLLM architecture is shown in Figure~\ref{fig:overall} and displays the base U-Net model interconnected with the LLaMA 3.2 Vision multimodal large language model (MLLM) through customized U-Net fusion and Cross-Attention fusion blocks. The U-Net encoder architecture incorporates ResNet50 (Residual Network) \citep{He_2016_CVPR} modules for the first four convolutional layers and the bridge connection, with the initial layers each outputting a skip connection. Each ResNet block encompasses three sets of convolutional neural network (CNN) blocks that are then summed to the original CNN input of each block through a residual connection. These CNN layers are then followed by a Batch Normalization layer and a Rectified Linear Unit (ReLU) nonlinear activation before passing to the next encoder stage. Leveraging the pretrained CNN layers into the encoder architecture through transfer learning is meant to enhance the model's generalization capabilities, speed training, and reduce overfitting. These skip connections are intercepted and passed through the proposed modules, where they are fused with key layers along the MLLM pipeline.

The MLLM architecture is composed of a vision encoder and a language decoder. The language decoder contains 40 sequential layers that consist of 32 self-attention blocks and 8 cross-attention blocks, which are inserted at layers 3, 8, 13, 18, 23, 28, 33, and 38. These cross-attention blocks serve to incorporate the vision encoder output into the language pipeline. Given that skip connections represent a vision component of the overall fusion architecture, cross-modal information fusion was implemented at cross-attention layers, specifically at layers 3, 18, 28, and 38 for each of the four U-Net encoder outputs. For incorporating the U-Net information into the MLLM, each skip connection was fused with the output of the MLLM vision encoder just prior to inclusion into each respective cross-attention layer, where it will proceed through the remaining language pipeline.

To evaluate the effectiveness of MDF-MLLM, input images were subjected to several preprocessing steps (detailed below) for standardization and then trained across several configurations seeking to optimize diagnosis accuracy. Model training was conducted on a series of publicly available retinal fundus images (FIVES, HRF, and StoneRounds datasets). A portion of the image dataset also includes anonymized patient information utilized by clinicians for health assessments. Together, these features were coordinated to optimize the MLLM’s ability to effectively categorize the patient’s health status.  

\subsection{Image Preprocessing and Prompt Initialization}
Retinal fundus images were presented to the model predominantly in native form with minimal preprocessing. This was intended to preserve the native features that will be made available to the MLLM. The only preliminary modifications involved resizing to $3 \times 512 \times 512$ ($C\times H\times W$) and normalization, which involved element-wise division by 255. Certain images also contained watermark identifiers from the original publishers. Augmentation steps were applied to remove these logos, and included watermark region isolation, conversion of the selected region to grayscale, threshold application to create a binary mask, expanding the mask with a $3 \times 3$ dilation kernel, mask removal from the original image region, and finally filling in the subtracted pixel regions using a Fast Marching Method (FMM) inpainting algorithm which propagates information from the surrounding pixels that remain. These steps serve as a means to simply standardize the visual inputs.

\begin{figure*}[h]
  \centering
  \includegraphics[width=\linewidth]{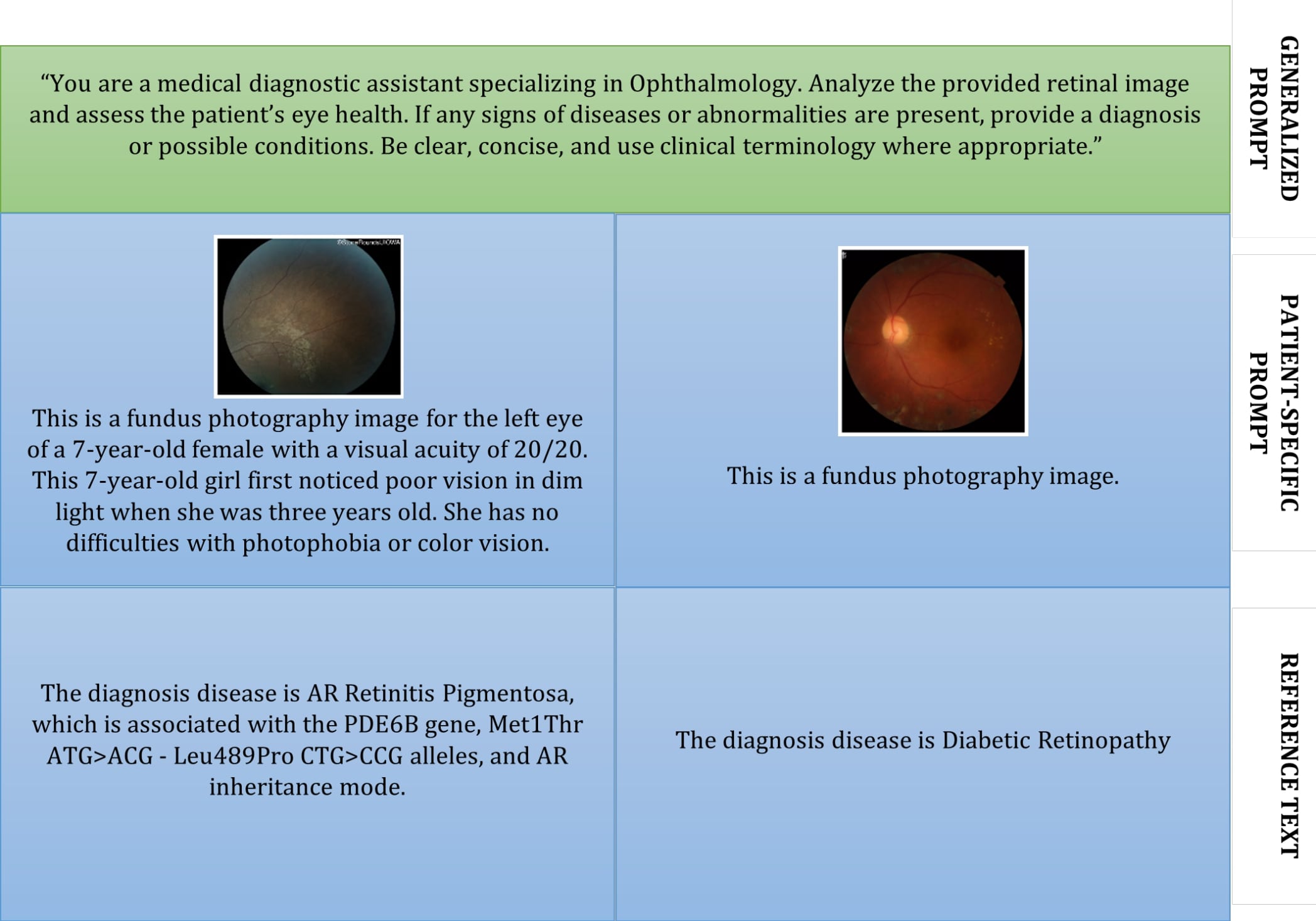}
  \caption{MLLM prompt layout with generalized prompt used on all data inputs (top), retinal fundus image with patient-specific information (center row), and reference text (bottom row) for model output comparison. The left column represents a patient with an inherited disease type, while the right column represents a patient with an acquired disease type.}
  \label{fig:prompt}
\end{figure*}

The MLLM architecture was fine-tuned to recognize features associated with varying disease types and to assess whether a specific fundus image belonged to a given disease class. To ensure consistent results, an instruction prompt was customized to guide the MLLM to act as an expert clinician and assess the health of a patient’s eye based on the provided fundus image and additional information, if available. Figure~\ref{fig:prompt} displays the full prompt used to train all images, along with examples of relevant patient information and fundus images with their respective ground truth labels. It should be noted that the images utilized from the public segmentation datasets (FIVES and HRF) did not contain additional patient information, other than disease classification. Therefore, these images were simply prompted with the phrase “This is a fundus photography image,” and the reference text only listed the specific disease type.

\subsection{Cross-Attention Fusion}
To fully integrate U-Net skip features into the MLLM pipeline, they were treated as vision components and fused to the MLLM vision encoder outputs before integrating into the model's cross-attention layers, which are dedicated to vision-language integration. For fusion implementation, we define $k$ skip feature maps as $S^{\left(k\right)}\in\mathbb{R}^{C_k\times H_k\times W_k}$ that have $C_k$ channels and spatial resolution $H_k\times W_k$. The feature map is divided into $p\times p$ non-overlapping patches ($16 \times 16$ for the present study), then flattened into vectors as shown below, with $N_k$ representing the number of patches.
\begin{equation}
\label{PK}
P^{(k)} = \mathrm{reshape}(\mathrm{unfold}_{p,p}(S^{(k)})) \in \mathbb{R}^{N_k \times C_k p^2}    
\end{equation}
\begin{equation}
\label{Nk}
N_k = \frac{H_k W_k}{p^2}
\end{equation}

Each patch vector is projected into the LLM embedding dimension $d$, which is 4096 on the LLaMA 3.2 11B Vision model, using a level-specific linear projection as shown below, with trainable weights ($W_{pat}^{\left(k\right)}$) and bias ($b_{pat}^{\left(k\right)}$).
\begin{equation}
\label{EK}
E^{\left(k\right)}=P^{\left(k\right)}W_{pat}^{\left(k\right)}+b_{pat}^{\left(k\right)}\in \mathbb{R}^{N_k\times d}
\end{equation}

To demonstrate cross-attention fusion, let $V_L\in\mathbb{R}^{T\times d}$ be the MLLM embedded vision features, where $T$ is the sequence length (which represents the number of patches on the MLLM vision encoder), and $d$ = 4096 is the hidden dimension of the MLLM embedded space. The MLLM vision features are fused with the U-Net encoder features using attention mechanisms by setting the query ($Q$), key ($K$), and value ($V$) parameters as shown below, with $|.|_2$ representing the Euclidian norm and $W$ and $b$ representing trainable parameters of a linear projection for each respective parameter $Q$, $K$, and $V$.
\begin{equation}
\label{QK}
Q^{\left(k\right)}=\frac{V_L}{|V_L|_2}W_Q^{\left(k\right)}+b_Q^{\left(k\right)}
\end{equation}
\begin{equation}
\label{KK}
K^{\left(k\right)}=\frac{E^{\left(k\right)}}{|E^{\left(k\right)}|_2}W_K^{\left(k\right)}+b_K^{\left(k\right)}
\end{equation}
\begin{equation}
\label{VK}
V^{\left(k\right)}=E^{\left(k\right)}W_V^{\left(k\right)}+b_V^{\left(k\right)}
\end{equation}

Finally, these projections were added to a vision transformer \citep{dosovitskiy2021imageworth16x16words} followed by a layer normalization step, as shown below, to better ensure compatibility with the MLLM pipeline. As previously described, $W_O$ and $b_O$ are trainable linear projection parameters, and $\alpha$ is a scaling constant controlling the strength of the injected fusion signal. Clamp is setting a range for the tensor such that all features above the range are set to the maximum value and all below the range are set to the minimum value. For the proposed implementation, $\alpha$ was set at 0.1 and the clamp range was set at (-5, 5).
\begin{equation}
\label{FK}
{\widetilde{F}}^{\left(k\right)}=\mathrm{softmax}\left(\frac{Q^{\left(k\right)}K^{\left(k\right)\top}}{\sqrt d}\right)V^{\left(k\right)}
\end{equation}
\begin{equation}
\label{FKK}
{\hat{F}}^{\left(k\right)}=\mathrm{LayerNorm}\left(\mathrm{Clamp}\left(\frac{{\widetilde{F}}^{\left(k\right)}}{|{\widetilde{F}}^{\left(k\right)}|_2}\cdot\alpha\right)W_O+b_O\right)
\end{equation}

\begin{figure*}[h]
  \centering
  \includegraphics[width=\linewidth]{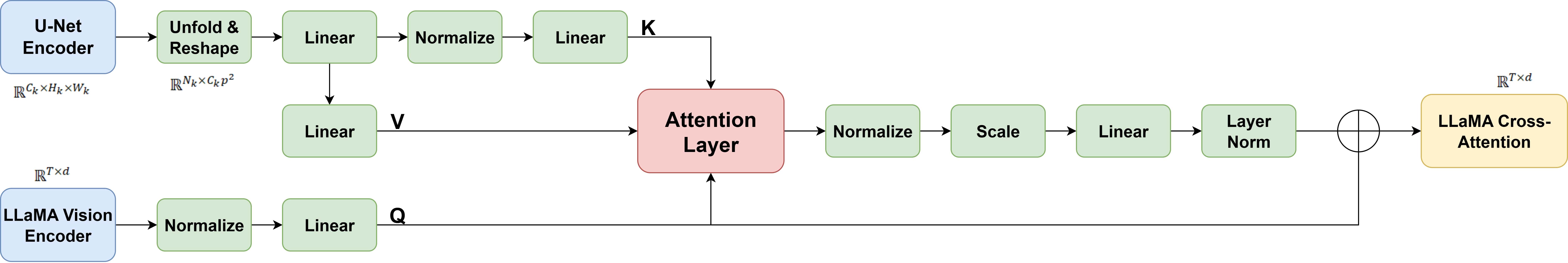}
  \caption{Cross-Attention Fusion block for merging U-Net skip connections to LLaMA Vision embeddings.}
  \label{fig:caf}
\end{figure*}

Upon completion of the vision embedding and skip feature fusion, the revised embeddings are summed to the original vision encoder parameters through a residual connection (shown below) and are passed as inputs to the MLLM cross-attention layer for integration with the language embeddings. Figure~\ref{fig:caf} provides a visual overview of the cross-attention fusion architecture.
\begin{equation}
\label{VLK}
{\hat{V}}_L^{\left(k\right)}=V_L+{\hat{F}}^{\left(k\right)}
\end{equation}

\subsection{U-Net Fusion}
Within the base U-Net encoder architecture, each layer includes a series of convolutional layers followed by Batch Normalization and a ReLU nonlinearity activation. This convolutional block is followed by a down-sampling layer that doubles the number of feature channels and halves the number of spatial features. These convolutional blocks are repeated three additional times, effectively reducing a $512 \times 512 \times 3$ input image to $64 \times 64 \times 1024$, and a copy of the output from each ReLU operation is saved as a skip connection. These connections would normally be concatenated to a corresponding up-convolution block within the decoder architecture of a traditional U-Net structure; however, the proposed fusion method seeks to incorporate additional vision and language feature information from the MLLM cross-attention layers before passing on to the decoder. With the primary objective of the present study being to enhance the MLLM classification capabilities, the purpose of this fusion module is solely intended to synchronize the gradient update mechanism within the MLLM architecture, which may otherwise have unintended effects on knowledge accumulation within the MLLM architecture during training.

Given that MLLMs are adept at learning global context features from input text and images, and the U-Net model is adept at learning more fine-grained semantic features, it is important to ensure that the higher-level context injected into the skip connection features is spatially well-aligned and matches channel depth. Otherwise, a portion of the U-Net's semantic understanding could be lost. To account for this, the U-Net fusion module incorporates the feature-wise linear modulation (FiLM) technique \citep{Perez_Strub_de_Vries_Dumoulin_Courville_2018}. MLLM embeddings ($H$) extracted from cross-attention layers are normalized to unit length and then mean-pooled across the sequence (axis $T$), as shown below. $H\in\mathbb{R}^{T\times d}$denotes the output of a cross-attention layer, where $T$ is the sequence length and $d$ = 4096 is the hidden dimension of the MLLM embedded space.
\begin{equation}
\label{barh}
\bar{h}=\frac{1}{T}\sum_{t=1}^{T}\frac{H_{t,:}}{|H_{t,:}|_2}\in\mathbb{R}^d
\end{equation}

The previous steps give a global descriptor vector per image that captures what the MLLM has ``understood'' about the `text + image' at a given layer in the architecture. It should also be noted that $T$, in this context, is different from the MLLM patches defined previously. Assuming  $S^{\left(k\right)}\in\mathbb{R}^{C_k\times H_k\times W_k}$ is the feature map output of the image from the $k$ level U-Net encoder, a channel-wise modulation vector is created for each skip map $k$, and an affine transformation is produced by:
\begin{equation}
\label{hatS}
{\hat{S}}_{c,y,x}^{\left(k\right)}=\gamma_c^{\left(k\right)}\cdot S_{c,y,x}^{\left(k\right)}+\beta_c^{\left(k\right)}
\end{equation}
where $S_{c,y,x}^{\left(k\right)}$ denotes the activation in spatial location $\left(y,x\right)$ in channel $c$, and $\gamma^{\left(k\right)}$ and $\beta^{\left(k\right)}$ (shown below) represent the linear projection slope and bias parameters, respectively.
\begin{equation}
\label{gammaK}
\gamma^{\left(k\right)}=\bar{h}W_\gamma^{\left(k\right)}+b_\gamma^{\left(k\right)}\in\mathbb{R}^{C_k}
\end{equation}
\begin{equation}
\label{betaK}
\beta^{\left(k\right)}=\bar{h}W_\beta^{\left(k\right)}+b_\beta^{\left(k\right)}\in\mathbb{R}^{C_k}
\end{equation}

From this, ${\hat{S}}^{\left(k\right)}$ will then be used as input to the decoder at level $k$. Binary segmentation masks included with the original images were used to calculate the U-Net loss as a part of the overall model gradient optimization.

\section{Results and Discussion}
\label{sec:expresults}
\subsection{Data Preparation and Evaluation}
\subsubsection{Datasets} 
Three primary data sources were used for this study. Two sources include the publicly available FIVES and HRF image datasets. FIVES (Fundus Image Vessel Segmentation) consists of 800 high-resolution fundus photographs with blood vessel segmentation masks manually annotated by trained medical personnel and verified by experienced ophthalmologists. The image set was collected from 573 patients and is further classified into labeled groups consisting of healthy patients and others diagnosed with one of three different disease types, including type 1 and type 2 diabetes mellitus, age-related macular degeneration, and glaucoma. The HRF (High Resolution Fundus) dataset contains 15 images of each healthy, diabetic retinopathy, and glaucomatous eye, and each with a corresponding manually annotated mask image.

The final dataset was extracted from the StoneRounds.org repository, which consists of a large database of de-identified clinical information obtained from individuals diagnosed with inherited retinal diseases with a known molecular cause at the Retina Clinic of the Department of Ophthalmology and Visual Sciences, University of Iowa \citep{STONE20171314}, and the NEI Data Commons \citep{neiDataCommonsWebsite} which consists of several NEI applications and datasets available to researchers. Many disease types are included in this repository; however, most would be considered exceptionally rare in their population prevalence. Therefore, only the top five disease classes, in terms of number of patients, within the dataset were selected for analysis. These include 490 images from 178 patients diagnosed with AR Retinitis Pigmentosa, AD Retinitis Pigmentosa, XL Retinitis Pigmentosa, Best Disease, or AR Stargardt Disease. In addition to retinal fundus images, many StoneRound patients also include clinical information that describes symptoms, family history, diagnosis reasoning, etc. This detailed information was helpful for fine-tuning the MLLM and is absent from the FIVES and HRF datasets. To complete the dataset, the selected StoneRounds images were professionally annotated for training with the U-Net architecture. Since the first two datasets lacked clinical text associated with the patient images, this research chose to exclude images associated with healthy patients since it represented only a single class and text is considered to contain key attributes to aid in model outputs and classification. Therefore, all results focused on distinguishing between diseased images grouped into either acquired (FIVES, HRF) or inherited (StoneRounds) disease types with the intent to expand the classification categories in future works.

\subsubsection{Evaluation Metrics}

\begin{figure}[h]
  \centering
  \includegraphics[width=\linewidth]{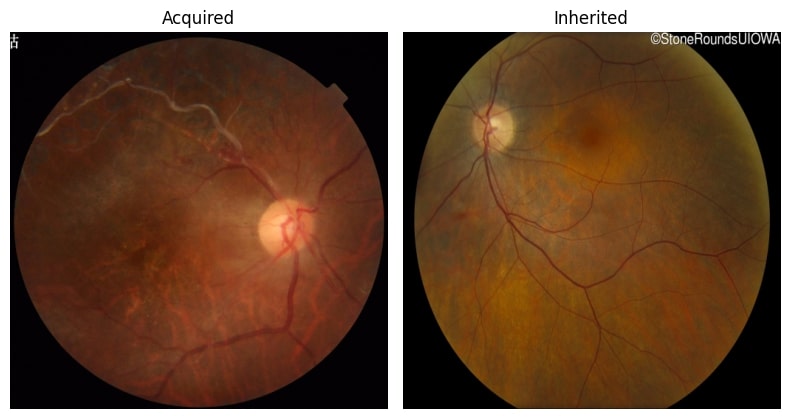}
  \caption{Example fundus images for each retinal disease classification category.}
  \label{fig:cat}
\end{figure}

To evaluate the performance of MDF-MLLM, classification accuracy, precision, recall, and F1-scores were chosen to assess the capability of the MLLM to identify the correct disease category. The dataset images were classified into one of two types that included either acquired diseases or inherited diseases. Acquired diseases are more common and represent abnormalities acquired over time as the patient ages or through some other affliction. In the present data, these types include glaucoma, diabetic retinopathy, and age-related macular degeneration. Inherited diseases represent abnormalities related to genetic disorders that were most likely acquired at birth. All disease types collected from the StoneRounds data were classified as inherited. Figure~\ref{fig:cat} displays example images for each category type. The primary motivation for generalizing disease classes considers the practical scenario of using such a model in a clinical setting. It would be more useful for an LLM to suggest a specific disease type only if it could also explain its reasoning for making such a suggestion, e.g., pointing out specific features that could be associated with the disease or other health factor information provided by the patient. In the current study, only the inherited disease types have additional patient information that could be provided as input to the model, making it difficult to learn enough information to make these types of suggestions. Therefore, without more meaningful data, it is likely that a clinician would be better served by having a model that could suggest a general disease type and prompt the need for additional diagnostic work.

\subsection{Implementation and Optimization}
The proposed MDF-MLLM architecture was developed and trained on a system of 3 nodes containing a total of 12 GPUs. Nodes 1 and 2 each included 4 NVIDIA L40S GPUs and a 64-core Intel Xeon Platinum 8562Y CPU. Node 3 included 4 NVIDIA A40 GPUs and a 32-core Intel Xeon Gold 6326 CPU. The overall model was developed in PyTorch and an AdamW optimizer was selected to adjust the parameters on both the U-Net and MLLM architectures during training. The initial learning rate was set to 1E-6, and the MLLM incorporated a StepLR scheduler with step size = 1 and gamma = 0.85. This scheduler was selected to better control the learning rate for fine-tuning the much larger MLLM model. During training, each experimental setup was trained over 60 epochs with a batch size of 2, and a reserved test dataset was used to select the best checkpoint epoch for validation. The full dataset of 1305 image-text pairs was split into train, validation, and test sets composed of 70\%, 10\%, and 20\% of the total samples, respectively, for evaluation.

For model optimization, a binary cross-entropy function (Equation~\ref{eq:lseg}) was chosen to calculate the segmentation loss on the U-Net output, while a cross-entropy function (Equation~\ref{eq:lmllm}) was utilized as the loss function on the generation of MLLM tokens. Equation~\ref{eq:ltot} shows the total combined loss on MDF-MLLM.
\begin{equation}
    \label{eq:lseg}
    Loss_{Seg} = -\frac{1}{N}\sum_{i=1}^N \big[ y_i \log(p_i) + (1-y_i)\log(1-p_i) \big]
\end{equation}

\begin{equation}
    \label{eq:lmllm}
    Loss_{MLLM} = -\sum_{k=1}^{K} y_k \log(p_k)
\end{equation}

\begin{equation}
    \label{eq:ltot}
    Loss_{total} = \frac{Loss_{Seg} + Loss_{MLLM}}{2}
\end{equation}

\subsection{Ablation Study and Benchmark Comparison}
The MLLM model was initially trained independently in order to construct a baseline performance with which the fusion model could be appropriately compared. The U-Net model was also independently pre-trained to ensure segmentation features were optimized prior to distribution to the MLLM cross-attention layers. These independent training steps revealed stark differences in learning speeds that would significantly affect the overall quality of a fused architecture design. Therefore, it was determined to utilize the pre-trained weights from these baseline models to initialize the MDF-MLLM architecture into a state of near convergence such that optimization of the fusion blocks would not become mis-aligned due to one model converging too quickly before the other. To evaluate this, our experimental setup involved loading the baseline weights into each model and fine-tuning the MDF-MLLM architecture in varying configurations with and without initialized weights frozen. Table~\ref{tab:scores} demonstrates the performance of MDF-MLLM, with comparison to the respective baseline MLLM.

The MLLM classification accuracy, recall, and F1-score improved considerably as a result of the U-Net fusion, which occurred with both U-Net and MLLM pretrained weights allowed to be further fine-tuned along with the fusion block parameters. This suggests that, although each individual model was independently optimized, it is critical that these parameters be further fine-tuned along with the parameters for the fusion modules to allow for the complete model to operate in a cohesive manner.

\begin{figure}[h]
  \centering
  \includegraphics[width=\linewidth]{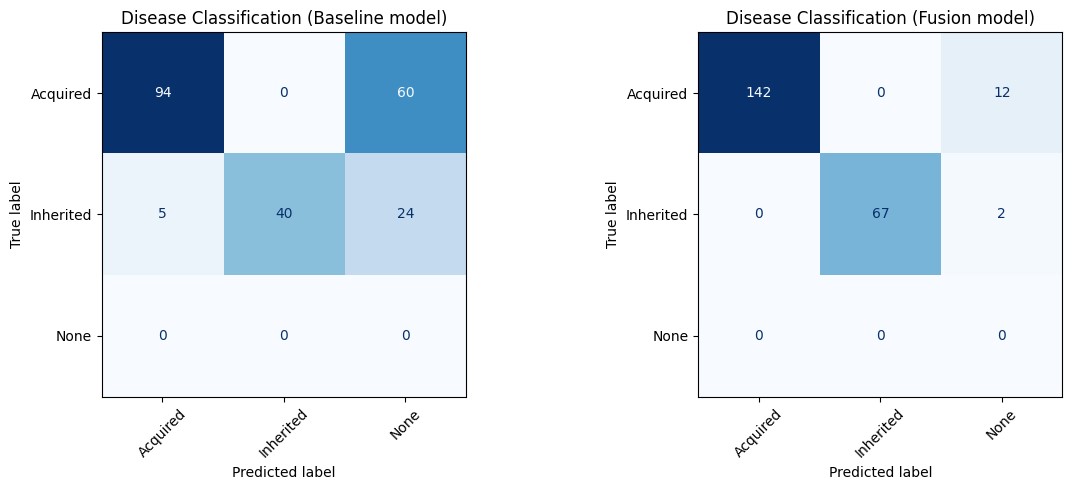}
  \caption{Confusion matrices demonstrating prediction capabilities for the baseline MLLM (left) and U-Net Fused MLLM (right). Sample images in which the MLLM did not mention either disease type in the generated text were classified as None.}
  \label{fig:mllm_conf_matrix}
\end{figure}

When classifying retinal fundus images into disease-related categories (inherited or acquired), MDF-MLLM was able to gain a significant level of understanding from the U-Net skip features, improving its classification accuracy by 56\% over baseline. Recall and F1-scores also improved by as much as 67\% and 35\% over baseline, respectively. All training scenarios had samples that were neither classified as acquired or inherited, but performed well among those that it did predict as either class, as indicated by the high precision metrics. However, the improvements in recall and F1-scores demonstrate how the fusion model was able to reduce the number of non-disease classifications. This is further highlighted in Figure~\ref{fig:mllm_conf_matrix}, which shows the confusion matrix classifications for both the baseline and fusion model. Sample images in which the generated MLLM text did not mention either disease type are shown as None. Incorporating skip features provided an enhancement of the low-level, fine-grained features of the MLLM vision encoder, enabling it to maintain richer spatial and semantic details, which align more effectively with textual representations via cross-attention. This architecture allows the model to gain a more expressive and compositional embedding space that results in improved classification decision boundaries.

As part of a visualization analysis, several text responses from each of the base MLLM and MDF-MLLM model outputs are shown in Table~\ref{tab:examples}. Each fundus image was provided with the general prompt specified in Figure~\ref{fig:prompt}, plus the additional patient-specific information listed in the first column. In general, the fusion and base models presented relatively coherent responses with the amount of information provided being reflective of the degree of detail provided in the prompt. 

For the classification categories of acquired diseases, patient information was not available to prompt the model, and therefore it only learned to provide a very generic response for these types of datasets. However, each model was not immune to a certain degree of hallucinations, as shown in the MDF-MLLM response in the last row of Table~\ref{tab:examples}. To determine the classification performance metrics for each architecture, a match search was performed within the response text to determine whether one of the disease classes for each of the acquired or inherited categories was present. From empirical observations and because the MLLM was instruction-tuned, nearly every response included a form of text phrased as “The diagnosis disease is …” or having a slight textual variation, but semantically equivalent. Therefore, it was assumed that the presence of a particular disease class was truly reflective of the model’s classification prediction and not due to uncorrelated, random text generation.

It should be noted that despite the pretrained baseline weights being utilized during initialization, differences in training speeds were still observed between each sub-architecture of the MDF-MLLM. The results reported in Table~\ref{tab:scores} demonstrate performance with the MLLM component optimized, which occurred after 55 training epochs. However, optimized U-Net performance occurred after 18 training epochs in the MDF-MLLM architecture, where the classification accuracy of the validation set was had not reached its peak and therefore the model needed to be further trained. This is an important trade-off to consider for individuals seeking to optimize such a model for a specific application.

\section{Conclusion}
\label{sec:conclusion}
Retinal fundus photography evaluation is a critical diagnostic tool used in ophthalmic health care, and the industry is rapidly adopting technological advances that seek to improve the speed and quality of clinical diagnoses. This research established a novel model architecture (MDF-MLLM) that enhances the classification capability for retinal fundus images into one of two disease categories. MDF-MLLM incorporates global contextual learning collected from images and text information using the LLaMA 3.2 multimodal LLM and fuses it with low-level semantic image features provided by the U-Net architecture. This approach sought to extract embedding features from each architecture, which could be fused and transferred back through each model’s processing pipeline, effectively creating a cross-modal learning environment. Most notably, MDF-MLLM saw a significant increase of 56\% in classification accuracy compared to the base MLLM architecture on FIVES, HRF, and the Disease Atlas of StoneRounds.org benchmark ophthalmology datasets.

With the overarching focus of the current study prioritizing enhancement of MLLM’s classification capabilities, future work will consider further optimization opportunities for the MDF-MLLM and include effective retinal vessel segmentation of the original fundus photography images. Additionally, new techniques will be investigated to compensate for the difference in training speeds of the fused models, with emphasis on providing better alignment for overall optimization of the combined architecture.

\section*{Data availability statement}
The original contributions presented in the study are included in the article. Further inquiries can be directed to the corresponding author.

\section*{Author contributions}
JJ: Conceptualization, Data curation, Formal analysis, Investigation, Methodology, Software, Model Training, Validation, Visualization, Writing – original draft. 
MAL: Conceptualization, Data curation, Formal analysis, Investigation, Methodology, Visualization, Writing – original draft. 
PK: Data curation, Investigation, Supervision, Writing – review \& editing. 
M-LS: Investigation, Supervision, Writing – review \& editing.
S-SC: Investigation, Supervision, Writing – review \& editing.

\section*{Funding}
The author(s) declare that no financial support was received for the research and/or publication of this article.

\section*{Conflict of interest}
The authors declare that the research was conducted in the absence of any commercial or financial relationships that could be construed as a potential conflict of interest.

\section*{Generative AI statement}
The author(s) declare that no Generative AI was used in the creation of this manuscript. Any alternative text (alt text) provided alongside figures in this article has been generated by Frontiers with the support of artificial intelligence and reasonable efforts have been made to ensure accuracy, including review by the authors wherever possible. If you identify any issues, please contact us.

\bibliographystyle{Frontiers-Vancouver}
\bibliography{arXiv}

\begin{table*}[ht]
\centering
\caption{MLLM and Fusion model performance by parameter initialization and disease type.}
\label{tab:scores}
\renewcommand{\arraystretch}{1.6}
\begin{tabular}{@{}l l c l c c c@{}}
\toprule
\textbf{Model Type} & \textbf{Parameter Initialization} & \textbf{Accuracy} & \textbf{Disease Type} & \textbf{Precision} & \textbf{Recall} & \textbf{F1-Score} \\
\midrule

\multirow{2}{*}{\textbf{Baseline MLLM}} & \multirow{2}{*}{MLLM unfrozen} & \multirow{2}{*}{0.60} & Acquired & 0.95 & 0.61 & 0.74 \\
 &  &  & Inherited & 1.00 & 0.58 & 0.73 \\

\midrule

\multirow{2}{*}{\textbf{Fusion}} & \multirow{2}{*}{\makecell[l]{U-Net frozen\\MLLM frozen}} & \multirow{2}{*}{0.37} & Acquired & 1.00 & 0.47 & 0.64 \\
 &  &  & Inherited & 1.00 & 0.12 & 0.22 \\

\midrule

\multirow{2}{*}{\textbf{Fusion}} & \multirow{2}{*}{\makecell[l]{U-Net frozen\\MLLM unfrozen}} & \multirow{2}{*}{0.78} & Acquired & 1.00 & 0.74 & 0.85 \\
 &  &  & Inherited & 1.00 & 0.88 & 0.93 \\

\midrule

\multirow{2}{*}{\textbf{Fusion}} & \multirow{2}{*}{\makecell[l]{U-Net unfrozen\\MLLM unfrozen}} & \multirow{2}{*}{\textbf{0.94}} & Acquired & 1.00 & \textbf{0.92} & \textbf{0.96} \\
 &  &  & Inherited & 1.00 & \textbf{0.97} & \textbf{0.99} \\

\bottomrule
\end{tabular}
\end{table*}

\begin{table*}[ht]
\setlength{\parindent}{0pt}
\caption{Example responses from the Fused and Base MLLM models, along with patient-specific prompt information and reference classification categories. Text in parentheses represents the true disease classifications. ARMD refers to Age-related Macular Degeneration and RP refers to Retinitis Pigmentosa.}
\label{tab:examples}
\renewcommand{\arraystretch}{1.5}
\begin{tabularx}{\linewidth}{|p{3cm}|X|X|p{2cm}|}
\hline
\textbf{Patient-Specific Prompt} & \textbf{Fusion MLLM Response} & \textbf{Base MLLM Response} & \textbf{Reference Category} \\
\hline
This is a fundus photography image. & The diagnosis disease is Glaucoma. & Diagnosis disease Glaucoma. & Acquired (Glaucoma) \\
\hline
This is a fundus photography image. & The diagnosis disease is Glaucoma. Prevention and Treatment Disease Glaucoma is a common chronic disease caused by high intraocular pressure, which can lead to optic nerve damage and vision loss. The diagnosis disease is Glaucoma. & The diagnosis disease is Glaucoma. & Acquired (ARMD) \\
\hline
This is a fundus photography image for the right eye of a 41 year old female with a visual acuity of 10/180. & The diagnosis disease is XL Retinitis Pigmentosa, which is associated with the RPGR gene, Glu1031 del1GA alleles, and XL inheritance mode. The clinical features supporting the diagnosis of XL Retinitis Pigmentosa in this patient include: XL inheritance mode, loss of acuity, night blindness, and fundus appearance showing XL Retinitis Pigmentosa. & The diagnosis disease is AR Retinitis Pigmentosa, which is associated with the ZNF408 gene, Glu536Stop GAG$>$TAG alleles, and AR inheritance mode. & Inherited (XL RP) \\
\hline
This is a fundus photography image for the right eye of a 37 year old male with a visual acuity of 20/15 - 1 sc. & PhD, Southern Illinois University at Carbondale, and his lab (member: PhD). The diagnosis disease is AR Retinitis Pigmentosa, which is associated with the USH2A gene, IVS10-2 A$>$G, Cys3358Tyr TGC$>$TAC alleles, and AR inheritance mode. & The diagnosis disease is Best Disease, which is associated with the BEST1 gene, Thr307Ala ACG$>$GCG alleles, and AD inheritance mode. & Inherited (AR RP) \\
\hline
\end{tabularx}
\end{table*}

\end{document}